\documentclass[letterpaper, 10 pt, conference]{ieeeconf}  
\IEEEoverridecommandlockouts                              

\usepackage{graphics} 
\usepackage{epsfig} 

\usepackage{stfloats}
\usepackage{times} 
\usepackage{amsmath} 
\usepackage{amssymb}  
\usepackage[hang]{subfigure} 
\usepackage{cite}
\usepackage{siunitx}
\usepackage{soul}
\usepackage{color}
\usepackage{xcolor}
\usepackage{mathrsfs}
\usepackage[ruled,linesnumbered]{algorithm2e}
\usepackage[font=scriptsize]{caption}

\soulregister\ref7  
\soulregister\cite7 
\soulregister\eqref7
\soulregister\prettyref7

\usepackage{multirow}
\usepackage{algpseudocode}
\usepackage{makecell}
\usepackage{booktabs}
\usepackage{threeparttable}
\usepackage{hyperref}
\usepackage{endnotes}

\newcommand{\eg}{\textit{e}.\textit{g}.}

\bstctlcite{IEEEexample:BSTcontrol}

\usepackage{prettyref}
\newrefformat{fig}{Fig.~\ref{#1}}
\newrefformat{sec}{Sec.~\ref{#1}}
\newrefformat{tab}{Tab.~\ref{#1}}
\newrefformat{algo}{Algo.~\ref{#1}}
\newrefformat{eq}{(\ref{#1})}

\usepackage{color}

\title{\LARGE \bf
Whole-body Humanoid Robot Locomotion with Human Reference}

\author{Qiang Zhang$^{1,*}$, Peter Cui$^{2,*}$,  David Yan$^{2}$, Jingkai Sun$^{1}$, Yiqun Duan$^{3}$, Gang Han$^{2}$, Wen Zhao$^{2}$, \\ Weining Zhang$^{2}$, Yijie Guo$^{2}$, Arthur Zhang$^{2}$, Renjing Xu$^{1,\dagger}$}

\begin{document}

\maketitle

\footnotetext[1]{The authors are with The Hong Kong University of Science and Technology (Guangzhou), China. {$^{*}$ are equal contributors, $^{\dagger}$ is the corresponding author} {\tt\small  \{qzhang749, jsun444\}@connect.hkust-gz.edu.cn, renjingxu@hkust-gz.edu.cn}}
\footnotetext[2]{The authors are with PNDbotics. {\tt\small \{peter.cui, david.yan, arthur.zhang\}@pndbotics.com}}
\footnotetext[3]{The author is with Human Centric AI Centre, Australia Artificial Intelligence Institute, University of Technology Sydney 2007 Ultimo Australia. {\tt\small yiqun.duan@student.uts.edu.au}}

\thispagestyle{empty}
\pagestyle{empty}

\begin{abstract}

Recently, humanoid robots have made significant advances in their ability to perform challenging tasks due to the deployment of Reinforcement Learning (RL), however, the inherent complexity of humanoid robots, including the difficulty of designing complicated reward functions and training entire sophisticated systems, still poses a notable challenge.
To conquer these challenges, after many iterations and in-depth investigations, we have meticulously developed a full-size humanoid robot, ``Adam”, whose innovative structural design greatly improves the efficiency and effectiveness of the imitation learning process.
In addition, we have developed a novel imitation learning framework based on an adversarial motion prior, which applies not only to Adam but also to humanoid robots in general.
Using the framework, Adam can exhibit unprecedented human-like characteristics in locomotion tasks.
Our experimental results demonstrate that the proposed framework enables Adam to achieve human-comparable performance in complex locomotion tasks, marking the first time that human locomotion data has been used for imitation learning in a full-size humanoid robot.
For more video demonstrations, please visit our YouTube channel: \href{https://www.youtube.com/watch?v=7hK2ySYBa1I}{https://www.youtube.com/watch?v=7hK2ySYBa1I}
\end{abstract}

\section{Introduction}
\label{sec:intro}

In recent years, the field of humanoid robotics has received widespread attention, with numerous research institutes and companies releasing cutting-edge innovations and research results successively, signifying the rapid development and rise of the field.
Boston Dynamics' Atlas robot~\endnote{\href{https://bostondynamics.com/atlas/}{https://bostondynamics.com/atlas/}} has demonstrated parkour-level mobility; Tesla's Optimus~\cite{optimus} and Figure's humanoid robots~\endnote{\href{https://www.figure.ai/}{https://www.figure.ai/}} have learned from human data to perform complex desktop manipulation tasks; the bipedal robot Cassie~\cite{gong2019feedback}~\cite{li2024reinforcement}~\cite{ICRA2021_RL-Cassie-Walking} and its humanoid version Digit~\endnote{\href{https://agilityrobotics.com/robots}{https://agilityrobotics.com/robots}}, are powered by motors and successfully move across a variety of terrains; the renowned legged robot company Unitree released their humanoid robot product H1~\endnote{\href{https://www.unitree.com/h1/}{https://www.unitree.com/h1/}}; Apptronik~\endnote{\href{https://apptronik.com/}{https://apptronik.com/}} has developed a humanoid robot named Apollo, powered entirely by push rod electric motors; OpenAI~\endnote{\href{https://openai.com/}{https://openai.com/}}, esteemed in the field of general-purpose AI, has acquired the 1X robot company~\endnote{\href{https://www.1x.tech/}{https://www.1x.tech/}} and proposed a development plan for embodied intelligence. 
The above indicates that humanoid robotics is becoming one of the key directions for researchers and companies and that mastering the core technology of humanoid robotics is crucial to bridging the gap between digital general-purpose AI and tangible hardware.

\begin{figure}[t]
    \centering
    \includegraphics[width=0.90\columnwidth]{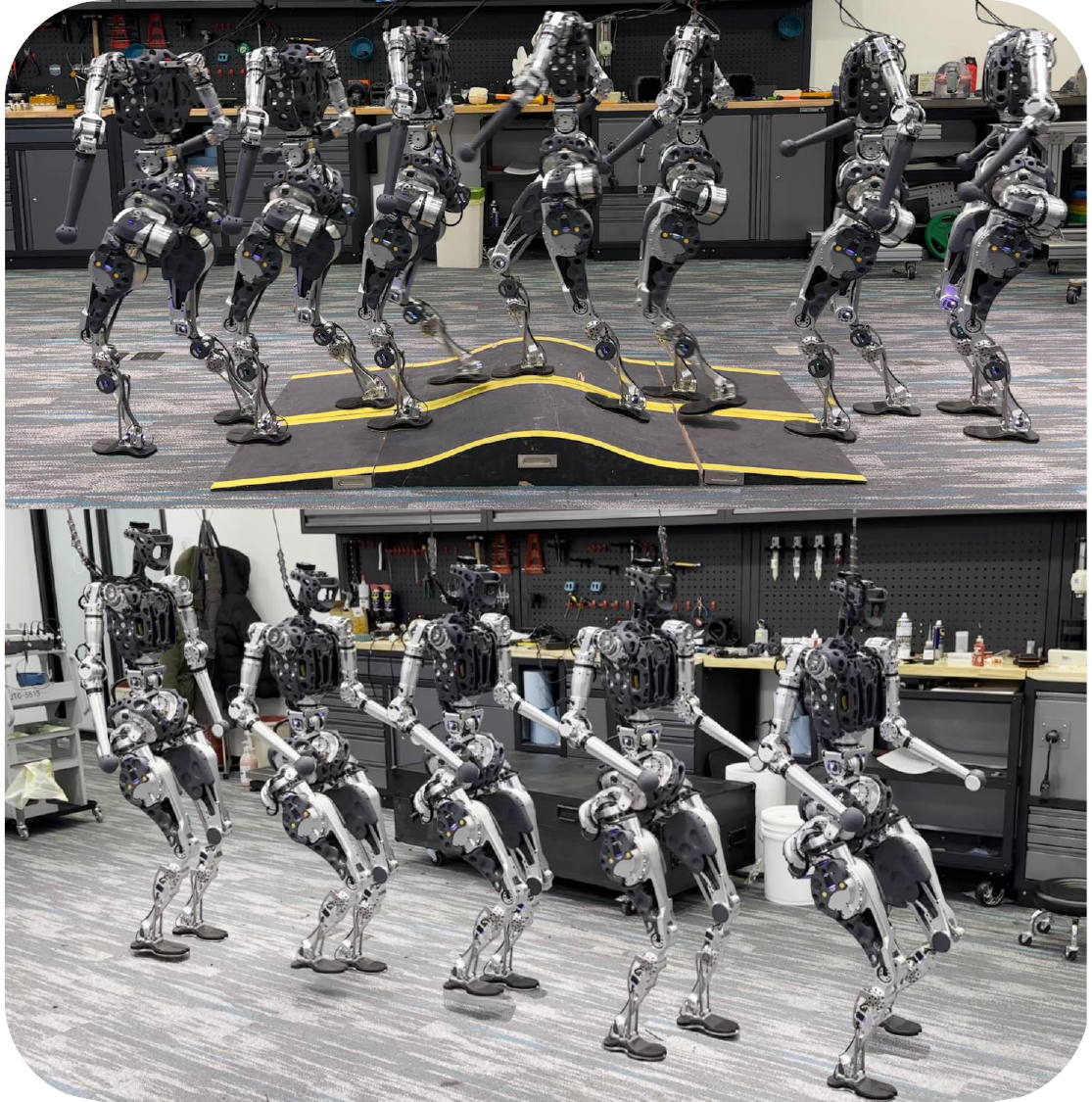}
    \caption{The top image displays the humanoid robot Adam walking on unseen terrain, and the bottom image shows Adam moving from a standing position to running.}
    \label{fig1}
\end{figure}
 
Traditional robot control algorithms, which typically rely on precise mathematical models and predefined motion planning, have proven to be very effective in the past for locomotion tasks of quadrupedal, bipedal, and humanoid robots.
Notably, Boston Dynamics' Atlas and Spot robots have demonstrated the effectiveness of these methods by using Model Predictive Control (MPC) algorithms~\cite{garcia1989model} to present impressive extreme mobility in a variety of demonstrations.
However, these algorithms often rely on accurate modeling of the environment, which can introduce significant challenges in terms of robustness and generalisability, especially in unknown or dynamically changing environments, where the performance of traditional control algorithms can be significantly degraded, limiting their usefulness in a wider range of application scenarios.
This reliance on accurate modeling, moreover, requires a high level of expertise to build and maintain these models, increasing the complexity of development and debugging.

Traditional robot control algorithms are notably limited in their adaptability, flexibility, and user-friendliness even though they exhibit excellent performance in specific environments, motivating researchers to explore alternative approaches to overcome these obstacles to design smarter and more adaptable robot control strategies.
Among them, deep neural network-based reinforcement learning algorithms have achieved promising results in the control of legged robots~\cite{rudin2022learning}. 
By interacting with the environment, Deep reinforcement learning algorithms are able to autonomously discover effective strategies to perform complex tasks and can be potentially extended to unknown or dynamically changing environments, which provides robots with unprecedented adaptability and flexibility.

Reinforcement learning algorithms have made significant progress in a variety of legged robots, but applications in the field of humanoid robotics still lack sufficient exploration.
This can be mainly attributed to the following factors: Firstly, most humanoid robots are expensive and difficult to maintain, which is a considerable obstacle for research institutes with limited funds; secondly, the interpretability problem of deep neural networks and the Sim2Real gap in the training process of deep reinforcement learning make it difficult to transfer the models to practical applications; Lastly, the complexity of humanoid robots greatly exceeds that of other legged robots, which makes the design of reward functions and training strategies during their training more challenging. 

To address these challenges, after repeated design and in-depth exploration, we have introduced the motor-joint-driven humanoid robot ``Adam", which has a significant cost advantage over traditional hydraulically-driven robots, and its modular design facilitates repairs during experiments and further reduces maintenance costs.
Moreover, our high-performance actuators ensure the robot's exceptional mobility, granting it a range of motion in its limbs close to that of a human. For the difficulty of setting up complicated reward functions in reinforcement learning, we adopt an innovative strategy of using human motion data to guide the learning process. Combined with our imitation learning training framework, Adam's performance on the locomotion task in the experiment is impressive.

In summary, we introduce a brand-new humanoid robot, Adam, and provide a new methodology and experimental validation for the learning, adaptation, and optimization of humanoid robots, paving a new way for research and development in humanoid robotics. The contributions of this paper can be summarized in the following three points:
\begin{enumerate}
\item We have developed and detailed an innovative biomimetic humanoid robot, Adam, whose limbs not only have a range of motion close to that of humans but also offer sufficient advantages in terms of low cost and ease of maintenance.
\item We designed and validated a new whole-body imitation learning framework for humanoid robots, which effectively solves the problem of complex reward function settings encountered in the reinforcement learning training of humanoid robots, greatly reduces the Sim2Real gap and improves the learning ability and adaptability of humanoid robots.
\item To address the Sim2Real challenges in complex humanoid robot reinforcement learning control algorithms, we incorporated numerous cross-validation and feedback adjustment steps into our framework. We not only demonstrated the robot's highly human-like performance in executing complex motion tasks but also provided a new perspective and data support for the future motion learning and optimization of humanoid robots.
\end{enumerate}


\section{Related work}
\label{sec:related}
\subsection{Legged robots locomotion}

Legged robot locomotion has seen significant advancement as it serves as the fundamental basis for achieving humanoid robotics. The incorporation~\cite{lirobust,siekmann2021blind} of Reinforcement Learning (RL) has played a crucial role in this field. The robot Cassie realizes a wide range of walking and running patterns utilizing periodic-parametrized reward functions~\cite{siekmann2021sim}, and even set a Guinness World Record for the fastest 100m dash using pre-optimized gaits~\cite{crowley2023optimizing}.
Jeon et al.~\cite{jeon2023benchmarking} underscores the effectiveness of potential-based reward shaping in expediting learning and enhancing the robustness of legged locomotion. Similarly, Shi et al.~\cite{shi2022reference} have broadened the capabilities of legged robots by introducing an assistive force curriculum that facilitates agile motion learning in settings without explicit references. The HRP-5P humanoid robot has demonstrated exceptional bipedal walking by leveraging feedback from actuator currents~\cite{singh2023learning}, while Kim et al.~\cite{kim2023torque} have proposed a torque-based approach to bridge the gap between simulated training and real-world application effectively.
In an innovative endeavor, DeepMind's research~\cite{haarnoja2023learning} enabled a miniature humanoid robot to master complex soccer skills through a unique teacher-student distillation and self-play methodology. Additionally, utilizing attention-based transformers in the Digit humanoid robot has facilitated more adaptable and versatile locomotion patterns~\cite{radosavovic2023learning}.
More recently, Tang et al.~\cite{tang2023humanmimic} employed adversarial critic component and specially designed Wasserstein distances to migrate locomotion from human reference. However, Adam has more flexibility and is more human-like overall, which leads to better performance in real robot experiments.

\subsection{Learning from Human Reference}
Humans, with their advanced intelligence and versatile locomotion capabilities, exhibit complex motion patterns that embody rich information. Leveraging insights from human behavior through learning from human references can greatly enhance the adaptability of robots. 
Traditional approaches to behavior cloning relied on manual programming, which proved to be time-consuming and inflexible~\cite{osa2018algorithmic,ravichandar2020recent}. Furthermore, defining the intricate and versatile locomotion skills of humanoid robots manually posed significant challenges~\cite{bohez2022imitate, han2023lifelike}. 

In recent years, imitation learning (IL) strategies have gained prominence, involving the tracking of either reference joint trajectories or extracted gait features~\cite{schaal1999imitation,van2010superhuman,jalali2019optimal,le2022survey}. However, these tracking techniques often operate on individual motion clips, resulting in discontinuities when transitioning between different locomotion patterns. 
To address this limitation, Generative Adversarial Imitation Learning (GAIL) was introduced by Peng et al., who proposed two innovative methods: AMP and Successor ASE~\cite{ho2016generative,peng2021amp,peng2022ase}. These approaches enable physics-based avatars to perform objective tasks while implicitly imitating diverse motion styles from extensive unstructured datasets. Variants of AMP have been successfully applied to learn agile quadrupedal locomotion and terrain-adaptive skills~\cite{escontrela2022adversarial, vollenweider2023advanced, li2023learning, wu2023learning, wang2023amp}. 
Additionally, Tang et al. introduced a Wasserstein adversarial imitation system with soft boundary constraints, further enhancing the capabilities of the AMP method \cite{tang2023humanmimic}. To facilitate the transfer of reference motion to robots, many works have introduced re-targeting techniques~\cite{ayusawa2017motion,grandia2023doc,tang2023humanmimic} that consider primitive skeleton and geometry consistency, enabling accurate dynamic modeling and complex balance controllers. However, AMP only makes robots learn kinematic motion relationships, which lacks physics constraints and causes a huge sim-to-real gap.

\vspace{-2mm}
\section{Preliminary}
\label{sec:preliminary}
\vspace{-1mm}
\subsection{The Structure of Humanoid Robot Adam}
\label{sec:robot_structure}

\begin{figure}
    \centering
    \includegraphics[width=1\linewidth]
    {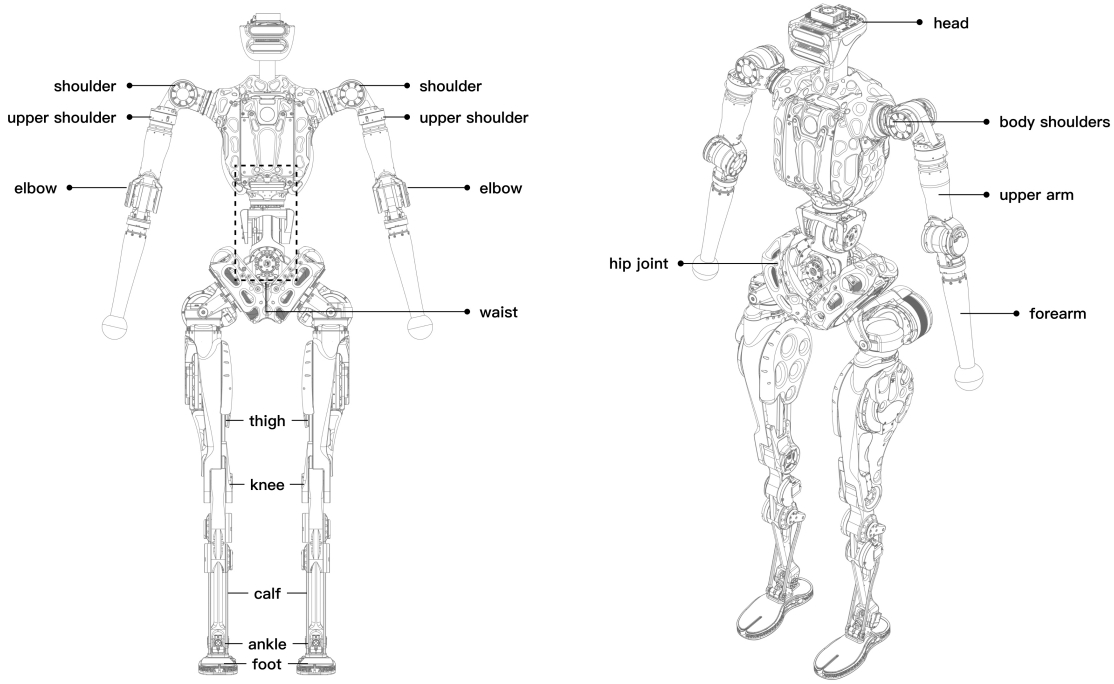}
    \caption{Schematic representation of the humanoid robot Adam(Lite), showing both the front view and the angled side view. Particularly, the design of its hip joint distinctly emulates the human skeleton, demonstrating a remarkable level of anthropomorphism throughout its structure.}
    \label{fig: robot}
\end{figure}

In this paper, we conducted experiments using the lite version of Adam.
Adam(Lite) is equipped with 25 QDD(quasi-direct drive), force-controlled PND actuators throughout its body, standing at 1.6 meters tall and weighing 60 kilograms. Its legs are fitted with four QDD high-sensitivity, highly back-drivable actuators with up to 340N$\cdot$m maximum torque. The arms have five degrees of freedom, and the waist has three. This fully modular, highly reusable design of flexible actuators, along with a highly biomimetic torso configuration, provides Adam with exceptional mobility and adaptability. The entire body utilizes a full-stack in-house design, real-time communication network PDN (PND Network), and PND actuators. The motion control computer is the 12th generation Intel i7 processor (Intel NUC) and PND RCU(Robot control unit). The PND RCU integrates all actuators, battery management system (BMS), power management, and is equipped with a 16-port Gigabit Ethernet programmable switch with network management capabilities, forming the robot’s sensory and control communication hub. This configuration enables Adam to perform large-scale parallel dynamic simulations and neural network training, achieving diversified full-body motion control that is truly applicable to real service scenarios, adapting to the complex and variable human social environment. Dexterous hands and vision modules can be optionally equipped. Since we focused on blind locomotion tasks in this paper, these parts were not included. 
The schematic diagram of Adam(Lite)'s entire body structure is shown in Figure.~\ref{fig: robot}, with information on its joint structure and more detailed in TABLE.~\ref{table:adam_specifications}.

\begin{table}[h!]
\centering
\caption{Specifications of Humanoid Robot Adam.}
\begin{tabular}{|c|c|}
\hline
Humanoid Robot & Adam (Lite) \\
\hline
Height & 1.6 M \\
\hline
Weight & 60 Kg \\
\hline
Full-body degrees of freedom & 25 \\
\hline
Single-leg degrees of freedom & Hip x3 + Knee x1 + Ankle x2 \\
\hline
Single-arm degrees of freedom & Shoulder x3 + Elbow x2 \\
\hline
Waist degrees of freedom & 3 \\
\hline
\end{tabular}
\label{table:adam_specifications}
\end{table}

\subsection{Motion Capture and Re-Targeting}

In our research, we explored a variety of human motion data sources to enrich our training set, ensuring that our model could learn diverse human motion characteristics. Initially, we utilized two public motion capture (mocap) databases: the SFU mocap dataset ~\endnote{\href{https://mocap.cs.sfu.ca/}{https://mocap.cs.sfu.ca/}} and the CMU mocap dataset ~\endnote{\href{http://mocap.cs.cmu.edu/}{http://mocap.cs.cmu.edu/}}.
Both datasets contain multiple motion capture sequences, covering a wide range of human activities, including everyday actions, sports movements, dance, and combat actions.
By integrating these two databases, we provided Adam with a diverse and high-quality human whole-body motion dataset, which is crucial for training the robot to understand and mimic human motion patterns.

Beyond public datasets, we also used high-precision motion capture equipment for custom motion recording. This approach allowed us to capture specific motion data, especially those special actions or sequences designed for particular experimental needs that are difficult to find in public databases. These custom motion data not only added diversity to our dataset but also enabled us to fine-tune and optimize our model more precisely, adapting it to specific motion tasks and challenges.
We did not approach re-targeting as an optimization problem, as was done in approaches like humanmimic~\cite{tang2023humanmimic}. Our objective was to secure high-quality motion data tailored for Adam, which led us to manually calibrate and convert each dataset ourselves.

\subsection{Reinforcement Learning on Legged Robots}
The application of RL to legged robots is modeled as a Partially Observable Markov Decision Process (POMDP), defined by the tuple $(\mathcal{S},\mathcal{O},\mathcal{A},\mathcal{R},p,\gamma)$ where $\mathcal{S}$ defines the state space, including all states of the environment. $\mathcal{O}$ denotes the partial observation space, representing a subset of the state space $\mathcal{S}$ that covers only the observable aspects of the environment by the agent. Concurrently, $\mathcal{A}$ specifies the action space, indicating all actions the agent is capable of executing. The reward function, denoted by $\mathcal{R}$, assigns a scalar reward to each state-action pair. The transition probabilities, represented by $p$, specify the likelihood of transition from the current state to a new state following the chosen action. The discount factor $\gamma \in [0,1]$, is applied to ensure that future rewards are appropriately weighted. 

Within this framework, at each timestep $t$, the agent observes $o_t \in \mathcal{O}$ from the environment. Based on this observation, the agent outputs an action $a_t \in \mathcal{A}$ sampled from a policy $\pi(a_t|o_t)$. Subsequently, the environment transitions to a new state $s_{t+1}$, governed by the probability distribution $s_{t+1} \sim p(s_{t+1}|s_t,a_t)$, and the agent is accordingly awarded a reward $r_t = \mathcal{R}(s_t,a_t)$. This objective function is formalized as to maximize the received reward:
\begin{equation}
\textnormal{arg}\max_{\theta} \mathbb{E}_{(s_t,a_t) \sim p_\theta(s_t,a_t)} \left[ \sum_{t=0}^{T-1} \gamma^t r_t\right]
\end{equation}
where T denotes the time horizon of POMDP.
\section{Method}
\label{sec:method}
\subsection{Learning with Human Reference by Adversarial Motion Priors}

\begin{figure}[t]
    \centering
    \includegraphics[width=0.9\linewidth]
    {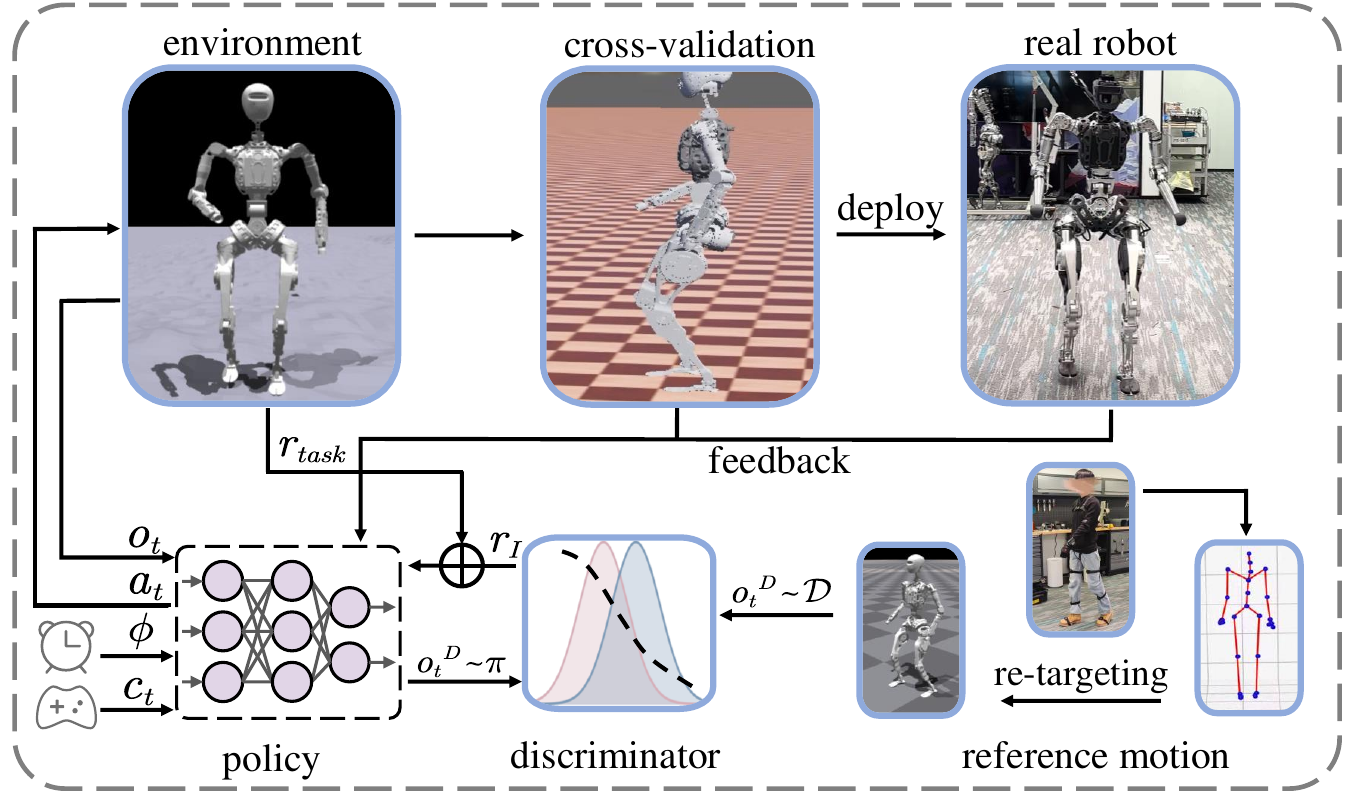}
    \caption{Adversarial Motion Priors Imitation Training Framework of Humanoid Robot}
    \label{fig: Framework}
\end{figure}
Our humanoid imitation framework is built upon the AMP. In our framework, the discriminator $D$ outputs the similarity between the state transitions sampled from the agent and it sampled from the reference demonstrations $\mathcal{D}$. It is crucial to select the discriminator observation $o_t^{D} \in \mathbb{R}^{58}$ fed into the discriminator to ensure the robots with similar state transitions can perform similar locomotion styles. The discriminator observation contains the velocity, position of each actuated joint and the positions of two hands and two feet of the humanoid. At each time step, we randomly sample the state transitions from the demonstrations and feed them into the discriminator to get the expert prediction loss for enabling the discriminator to distinguish them,
\begin{equation}
\mathcal{L}_{expert}=\mathbb{E}_{(o_t^{D}, o_{t+1}^{D})\sim\mathcal{D}}[(D(o_t^{D}, o_{t+1}^{D})-1)^2]
\end{equation}
As do the same with state transitions sampled from the policy,
\begin{equation}
\mathcal{L}_{policy}=\mathbb{E}_{(o_t^{D}, o_{t+1}^{D})\sim\pi}[(D(o_t^{D}, o_{t+1}^{D})+1)^2]
\end{equation}
We follow to~\cite{peng2021amp} to formulate the penalty for gradients on samples from the reference to stabilize training,
\begin{equation}
\mathcal{L}_{GP}=\mathbb{E}_{{(o_t^{D}, o_{t+1}^{D})\sim\mathcal{D}}}\left[\| \triangledown \mathcal{D}({o_t^{D}, o_{t+1}^{D}}) \|^2\right]
\end{equation}
Finally, we formulate the total AMP loss as,
\begin{equation}
\begin{aligned}
\mathcal{L}_{AMP} =\frac{1}{2}\mathcal{L}_{expert}+\frac{1}{2}{L}_{policy}+\lambda_{GP}\mathcal{L}_{GP}
\end{aligned}
\end{equation}
The AMP loss function directs the discriminator to rate samples, giving scores close to +1 for genuine reference motions and nearing -1 for those generated by the policy. The objective of the policy is to create motions convincing enough to lead the discriminator into assigning higher scores, demonstrating its ability to closely mimic reference motions. Subsequently, the formulation of the imitation reward for the policy's training is denoted as,
\begin{equation}
    r_I = \text{max}[0,1-\frac{1}{4}(D(o_t^{D}, o_{t+1}^{D})-1)^2]
\end{equation}
where $o_t^{D}, o_{t+1}^{D}$ are sampled from the policy.
\subsection{End-to-End Reinforcement Learning on Humanoid Robot}
Concurrently, the movement direction in reference motion is typically limited to the local coordinate. To facilitate control under the world coordinate system, generate more natural gaits, and enable a more effective transition from simulation to reality on challenging terrains, we introduced coordinated task rewards. The task rewards are composed of three parts, the command reward, the periodic reward, and the regularization reward. The command reward forces the robot to move alone in the command directions, which is formulated as,
\begin{equation}
\vspace{-1mm}
r_{com}=\sum\lambda_i\text{exp}(-\omega_i(|\mathbf{v^{i}_{des}}-\mathbf{v^{i}_{t}}|)\quad i \in(x,y,yaw)
\end{equation}
where $\lambda_i$ and $\omega_i$ are the weight of each direction of command rewards, $\mathbf{v}_{des}$ is the desire velocity vector along the specific direction and $\mathbf{v}_t$ is the velocity vector at $t$.

To facilitate the attainment of the desired gait performance, we introduce periodic rewards aligned with imitation rewards. This method naturally promotes the maintenance of a stable gait in the robot. However, if a variable gait is desired, it would be advisable to omit this reward function. We follow~\cite{siekmann2021sim} to formulate the periodic reward by the swing phases, characterized by the foot moving through the air, and the stance phases, where the foot is firmly planted on the ground. Each periodic reward item is composed of a coefficient $\alpha_i$, a phase indicator $I_i(\phi)$, a valued phase
reward function $V_i(s_t)$, $\phi$ is the cycle time, $i$ denotes the phase is stance phase or swing phase. The swing and stance phases are sequentially arranged and collectively span the entire cycle duration by establishing a ratio $\rho \in (0,1)$. This configuration ensures that the swing phase occupies a duration equivalent to $\rho$, followed immediately by the stance phase which extends for a duration of $1-\rho$. The single-foot reward is written as follows,  
\begin{figure}
    \centering
    \includegraphics[width=0.9\linewidth]
    {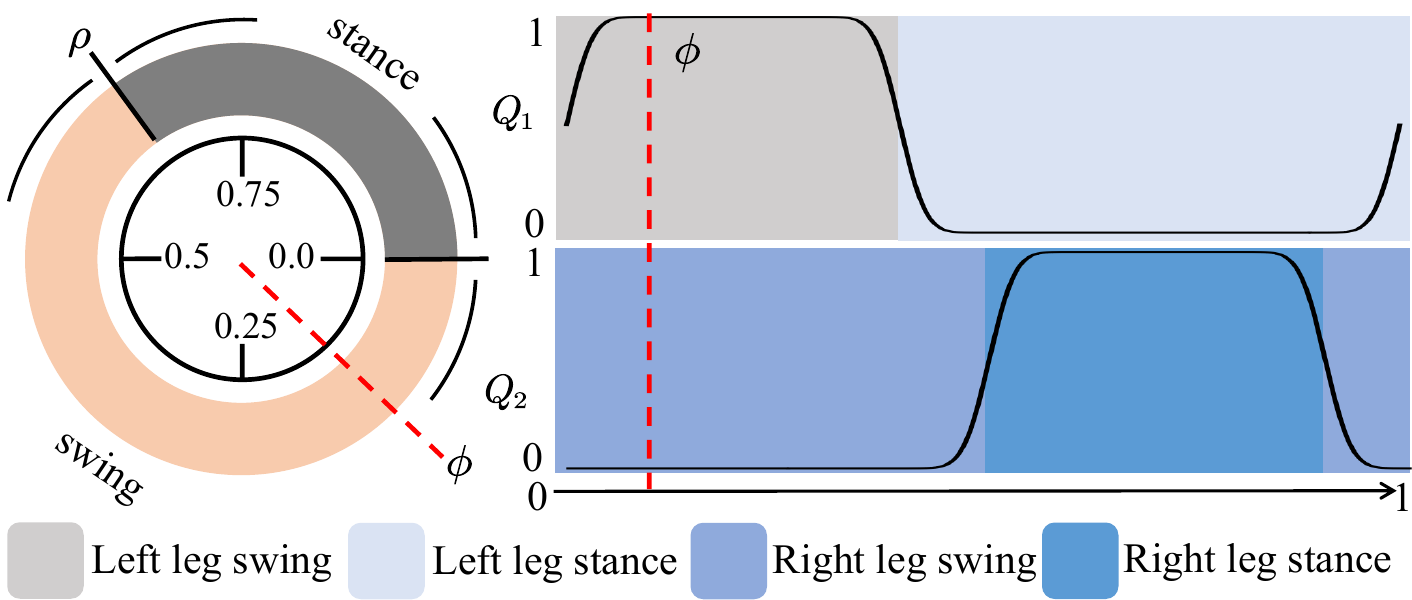}
    \caption{Visualization of periodic rewards based on the von Mises distribution}
    \vspace{-1mm}
    \label{fig: periodic}
\end{figure}
\begin{equation}
\begin{aligned}
&r_{per} = \sum \alpha_i \mathbb{E}[I_i(\phi)] V_i(s_t)\\
&V_{stance}(s_t) = \text{exp}(-10F_{f}^2)\\
&V_{swing}(s_t) =  \text{exp}(-200v_{f}^2)
\end{aligned}
\vspace{-1mm}
\end{equation}
where $F_{f}$ is the norm force of each foot, $v_{f}$ is the speed of each foot. For modeling the phase indicator $I_i(\phi)$, we follow the \cite{park2001impedance} to use the mathematical expectation of Von Mises distribution. The visualization of the phase indicator is shown in Figure~\ref{fig: periodic}. And we formulate as, 
\begin{equation}
\begin{aligned}
    &Q_1=I_{stance}(\phi+\theta_{left}) \\
    &Q_2=I_{stance}(\phi+\theta_{right}) 
\end{aligned}
\vspace{-1mm}
\end{equation}
where the $\theta_{left}$, $\theta_{right}$ is the offsets of left and right leg in cycle time.
Expect the normal periodic reward, we calculate rewards for foot speed, height difference, and the symmetric during the swing phase for more natural gait style. The foot speed tracking reward is formulated as,
\begin{equation}
\begin{aligned}
&q^i=\text{clip}(\frac{\phi}{\rho} - 0.5,0,1)\\
&r(s_t) = \left\{ 
  \begin{array}{ll}
  16(q^iv_{f}^i)^2 & 0 \leq q_i \leq 0.6, \\
  0 & q_i > 0.6.
  \end{array} 
\right.
\end{aligned}
\vspace{-2mm}
\end{equation}
where $i \in (\text{left},\text{right})$, ``clip'' is the function to constrain the values of variables within the range of 0 to 1. The foot speed tracking reward encourages the robots to perform higher foot speed during the swing phase. The height difference reward is,
\begin{equation}
\begin{aligned}
&q^i=\frac{\phi}{\rho}\\
&\delta h = h^i_{f}-h^{-i}_{f} - 0.02\\
&r(s_t) = \left\{ 
  \begin{array}{ll}
  2\text{exp}(-25(|\delta h|) & 0 \leq q_i \leq 0.3, \\
  0 & q_i > 0.3.
  \end{array} 
\right.
\end{aligned}
\vspace{-2mm}
\end{equation}
where $h^i_{f}$ is the height $i$ foot, the $h^{-i}_{f}$ denotes the height of the other side foot. The purpose of this function is to calculate reward based on the height difference of the foot only during certain early stages of the gait cycle. The symmetric reward is written as follows,
\begin{equation}
\begin{aligned}
&\mathbf{d}_t=\mathbf{p}_t^{left} - \mathbf{p}_t^{right}\\
&tf=(\mathbb{E}[I_{left}(\phi)]  > 0.5)\wedge(\mathbb{E}[I_{right}(\phi)]  > 0.5)\\
&\delta \mathbf{f}_t = tf \cdot \mathbf{d}_t + \neg tf \cdot \delta \mathbf{f}_{t-1}\\
&\delta \mathbf{l}_{t}= \neg tf \cdot \delta \mathbf{f}_t + tf \cdot \mathbf{d}_t\\
&r(s_t) =3.3tf\text{exp}(-10||\mathbf{d}_t + \delta \mathbf{l}_{t}||_1)
\end{aligned}
\vspace{-2mm}
\end{equation}
where $\mathbf{p}_t^{i}$ is the 3-D position of foot end-effector, $\neg$ is negation symbol. This function calculates a reward based on the symmetry of the gait swing, where the gait phases of the left and right feet are taken into consideration, along with the distance between the feet. 

To enhance the robustness of the sim-to-real transfer, we incorporated regularization rewards into the comprehensive reward structure. These rewards enforce motion constraints, emphasizing both smoothness and safety. The detailed formulation of each reward is shown in Table~\ref{tab: regularization rewards}. The DoF limits reward and keeps the movement within the robot's physical capabilities, ensuring that it doesn't attempt movements that could damage its mechanisms or be outside its operational scope. $\mathbf{b}$ denotes the vector angle of DoF. $\mathbf{b}_{lower}$ and $\mathbf{b}_{upper}$ are the upper and lower bounds of the joint limit. The DoF velocity reward forces the system to maintain optimal speeds. Similarly, the DoF acceleration reward encourages smooth increases and decreases in speed, contributing to the overall stability. For robotic arms, the arm DoF penalty is applied to discourage large positions or movements. The torso yaw reward and orientation differential rewards are specific to the rotational movement around a robot's vertical axis. This control is vital for balance and orientation, especially in humanoid robots.
\begin{table}[]
    \centering
    \caption{Regularization Rewards}
    \scalebox{1}{
\begin{tabular}{ll}
\hline
Item                     & Detail \\ \hline
Action differential      & $\text{exp}(-0.05\|\mathbf{a}_t - \mathbf{a}_{t-1}\|_2)$   \\
DoF limits               & $\text{exp}(-2.0(\text{min}(0,\mathbf{b} - \mathbf{b}_{upper})$\\
                         &$-\text{max}(0,\mathbf{b}-\mathbf{b}_{lower}))) $  \\
DoF velocity             & $\text{exp}(-1 \times 10^{-4}\|\dot{\mathbf{b}}\|^2_2)$   \\
DoF acceleration         & $\text{exp}(-1 \times 10^{-7}\|\ddot{\mathbf{b}}\|^2_2)$   \\
Arm DoF penalty          & $\text{exp}(\|\mathbf{b}_{arm}\|_1)$   \\
Orientation differential & $\text{exp}(-300(roll^2+pitch^2))$   \\
Torso yaw                & $\|b_{torso~yaw}\|_1$   \\ 
Torques                  & $\text{exp}(-5 \times 10^{-4}\|\mathbf{\tau}_t\|_2)$    \\\hline
\end{tabular}
    }
    \vspace{-2mm}
    \label{tab: regularization rewards}
\end{table}
\begin{table}[]
    \centering
    \caption{Domain Randomization Range}
    \scalebox{1}{
\begin{tabular}{lll}
\hline
Randomization Item     & Range            & Unit \\ \hline
Mass                   & {[}-0.05,0.05{]} & Kg   \\
Center of mass alone x & {[}-0.05,0.05{]} & m    \\
Center of mass alone z & {[}-0.05,0.05{]} & m    \\
Motor strength         & {[}0.7,1.4{]}$\times$default    & N$\cdot$m   \\
Impulse                & {[}0,0.8{]}      & m/s  \\
External force         & {[}-500,500{]}   & N    \\
Linear velocity        & {[}0.8,1.2{]}$\times$default    & m/s  \\ \hline
\end{tabular}
    }
    \label{tab: domain randomization}
\end{table}

\vspace{-0.5mm}
\section{Experiments}
\vspace{-0.5mm}
\label{sec:experiments}

\begin{figure*}[htbp]
\centering
\includegraphics[width=0.9\textwidth]{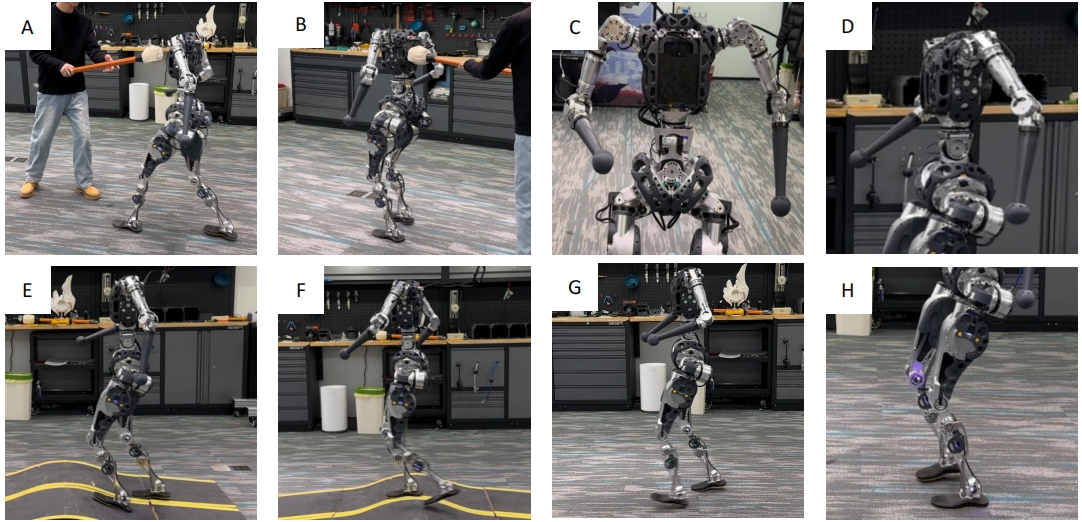}
\caption{\textbf{Real Robot Experiments.} We tested our method on Adam. (A)(B) demonstrate the robot's robust locomotion performance under external disturbances. \textbf{Especially, it demonstrated the characteristic of ``straight-knee''} (C)(D) show the human-like swinging of the upper limbs. (E)(F) display the robustness and human-likeness on unknown and complex terrains. \textbf{(G)(H) present for the first time the humanoid robot's ``heel-to-toe'' running and walking gaits.}}
\vspace{-6mm}
\label{fig: demo}
\end{figure*}

\subsection{Training and Implementation Details}
\subsubsection{Training Setup} 
\begin{figure}[t]
    \centering
    \includegraphics[width=0.9\linewidth]
    {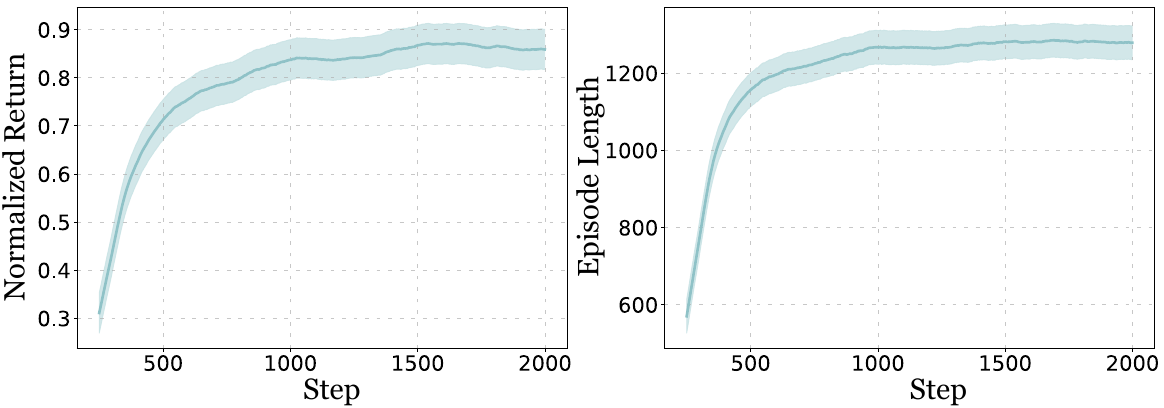}
    \vspace{-2mm}
    \caption{Curve of normalized return and episode length of policy training}
    \vspace{-4mm}
    \label{fig: curve}
\end{figure}

We trained our policy via a model-free reinforcement learning algorithm known as Proximal Policy Optimization (PPO)~\cite{schulman2017proximal} on 4096 Isaac Gym simulation environments parallel. 
In humanoid locomotion scenarios, actions are represented by $a_t \in \mathbb{R}^{25}$ a 25-dimensional vector specifying the desired positional adjustments for each actuated joint as dictated by the Proportional-Derivative (PD) controller. Observations, denoted as $o_t \in \mathbb{R}^{91}$, encompass the humanoid's current linear and angular velocities, the average velocities in the $(x,y, yaw)$ directions, and the orientation of the gravity vector within the robot's base frame, all measured by an inertial measurement unit (IMU). The position and velocity of each joint are captured by the actuators' encoders. A command, $c_t = (v^{x}_{des},v^{y}_{des},\omega^{yaw}_{des})$, is introduced to specify the desired velocities along the x-axis, y-axis, and yaw within the robot's base frame. Periodic observations include the sine and cosine of the cycle time and the swing phase ratio $\rho$, to dynamically adjust the gait. Additionally, the root's height is incorporated into the observations to manage the humanoid robot's posture. For enhanced performance, the action executed in the previous timestep is also considered.
Our framework ensures a smooth training process for Adam, as illustrated in Figure~\ref{fig: curve}.
\begin{figure}[t]
    \centering
    \includegraphics[width=0.9\linewidth]
    {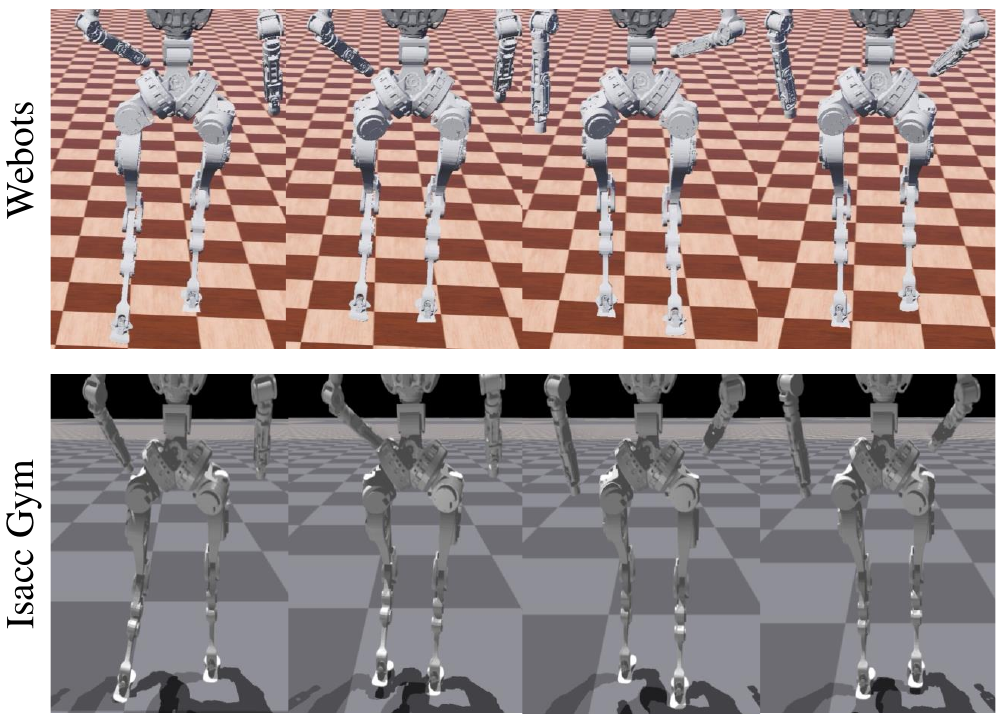}
    \caption{Cross-validation of Webots and Isaac gym simulator in forward locomotion scenario}
    \label{fig: cross-validation}
    \vspace{-2mm}
\end{figure}
\subsubsection{Domian Randomization}
This randomization strategy aims to tackle three critical uncertainties: modeling discrepancies in the robot and its environment, sensor noise, and unobservable disturbances and impulses. To avoid the sim-to-real gap in force applying in Isaac Gym, we use the sudden velocity on the root to replace the force impulse. These dynamic properties are represented in Table~\ref{tab: domain randomization}, with values adjusted across the specified ranges. The mass of individual robot bodies (\eg, torso), is varied to simulate different weight distributions. The center of mass is randomly shifted in specific directions. These techniques help the policy to adapt to the uncertainties in robot components. Motor strength influences the torque output to deal with the uncertainties in motors. Furthermore, impulses are applied by adding velocities in the x-y plane on the robot root, replicating the effect of sudden external shocks. External forces are also directly applied to the body of the robot to simulate unobservable environmental forces. Additionally, noise is injected into the linear velocity to simulate speed estimation errors.
A series of tests were performed under a variety of conditions and in response to external disturbances, with the outcomes illustrated in Figure.~\ref{fig: demo}. These results shows our method provided a new perspective of future motion learning for humanoid robots
\vspace{-0.5mm}
\subsection{Cross-validation and Feedback Fine-tuning}
We carried out comprehensive cross-validation experiments using both Webots~\endnote{~\href{https://cyberbotics.com/}{https://cyberbotics.com/}} and Isaac Gym as simulation platforms to evaluate the effectiveness of our models in environments that closely mimic real-world conditions. Notably, Webots offers advanced physical simulation capabilities, enabling us to validate our policy across diverse platforms while fine-tuning hyperparameters for optimal performance. Which vividly demonstrated in Figure~\ref{fig: cross-validation}, where the humanoid robot's gait appears synchronized and uniform across both simulators, underscoring the model's reliability. Additionally, we conducted real-world tests to further validate our model's practical applicability. The seamless transition of the humanoid robot from a standing to a running state in the real world is captured in Figure~\ref{fig1}, highlighting the model's effectiveness and the potential for real-world deployment.

\section{Conclusion}
\label{sec:conclusion}

In this paper, we implement a whole-body imitation learning framework on the humanoid robot Adam, whose performance in complex gaits tasks is comparable to that of humans, and demonstrate human-like features such as ``heel-to-toe'' transitions and straight-knee movements for the first time in humanoid robots. Our experiments have demonstrated the practicality and efficiency of our framework, highlighting its great potential for further research in humanoid robotics. Moreover, our framework supports the integration of perceptual modules, and in the future, we plan to incorporate more sensors to allow Adam to imitate human motions under integrated perceptual conditions.

\bibliographystyle{IEEEtran}
\typeout{}
\bibliography{IEEEabrv,mybibfiles}
\theendnotes
\end{document}